\documentclass[11pt]{article}
\usepackage{amsmath}
\usepackage[final]{acl}

\usepackage{times}
\usepackage{latexsym}
\usepackage{amssymb} 
\usepackage{hyperref}
\usepackage[T1]{fontenc}

\usepackage[utf8]{inputenc}

\usepackage{microtype}

\usepackage{inconsolata}

\usepackage{graphicx}

\usepackage{booktabs}  
\usepackage{float}
\newcommand{\sensory}[1]{\texttt {#1}}
\newcommand{\hyp}[1]{\textbf{H$_{\text{#1}}$}}

\title{Words that make SENSE:\\Sensorimotor Norms in Learned Lexical Token Representations}

\author{Abhinav Gupta, Toben H. Mintz, Jesse Thomason \\
 University of Southern California\\
  \texttt{\{abhinavg, tmintz, jessetho\}@usc.edu} \\}

\begin{document}
\maketitle
\begin{abstract}

While word embeddings derive meaning from co-occurrence patterns, human language understanding is grounded in sensory and motor experience. 
We present \textsc{SENSE} (\textbf{S}ensorimotor \textbf{E}mbedding \textbf{N}orm \textbf{S}coring \textbf{E}ngine), a learned projection model that predicts Lancaster sensorimotor norms from word lexical embeddings.
We also conducted a behavioral study where 281 participants selected which among candidate nonce words evoked specific sensorimotor associations, finding statistically significant correlations between human selection rates and \textsc{SENSE} ratings across 6 of the 11 modalities.
Sublexical analysis of these nonce word selection rates revealed systematic phonesthemic patterns for the \sensory{interoceptive} norm, 
suggesting a path towards computationally proposing candidate phonesthemes from text data.

\end{abstract}

\section{Introduction}
Grounded cognition theory posits that humans rely on multimodal representations from perceptual, motor, and introspective experiences for cognitive activities~\cite{barsalou2008grounded}.
Empirical studies suggest that comprehending action, perceptual and abstract concepts elicits rapid, automatic activity in modality-specific brain areas, indicating a strong relationship between these concepts and human embodied experience~\cite{barsalou2008grounded, HAUK2004301, Vigliocco2014}.

The Lancaster sensorimotor norms dataset quantifies these sensorimotor 
associations by averaging ratings from 3,500 participants for 39,707 
English words across six perceptual modalities (\sensory{auditory}, 
\sensory{gustatory}, \sensory{visual}, \sensory{haptic}, \sensory{olfactory}, and \sensory{interoceptive}) and five action effectors (\sensory{hand/arm}, \sensory{foot/leg}, \sensory{head}, \sensory{torso}, and \sensory{mouth/throat})~\cite{Lynott:19}. 
Unlike the five traditional senses, \sensory{interoceptive} ratings capture internal bodily sensations including emotional and visceral experience, dimensions shown to be particularly relevant to the grounding of abstract concepts~\cite{Ponari2018, Vigliocco2014}.

Systematic form-meaning associations called phonesthemes are sublexical units carrying consistent associations across words, for example, ``gl-'' in glitter, gleam, and glow~\cite{Bergen2004}.
Sublexical processing is crucial for unfamiliar words and language learning~\cite{UBELLACKER2022211, INDEFREY2009517}, but whether neural language representations encode such correspondences is unclear~\cite{abramova2013automatic}.

Current language models rely on word embeddings learned from co-occurrence patterns, and are not explicitly trained to encode sensorimotor information.
Prior work has demonstrated that Lancaster norms, used either as standalone word embeddings or in combination with other embeddings, achieve competitive performance on tasks and benchmarks such as GLUE and the Visual Dialog task~\cite{Das_2019} when incorporated into a pre-trained language model~\cite{kennington-2021-enriching}. 
We aim to explore how much information from these sensorimotor norms is already encoded within existing word embeddings.
Studies have found correlations between word embeddings and human embodied, perceptual and introspective experiences~\cite{Utsumi2020, Louwerse2009, Lenci2018emotions, kennington-2021-enriching}.
However, most prior work has focused primarily on emotional information~\cite{Lenci2018emotions}, evaluated small word sets (e.g., approximately 500 words~\cite{Utsumi2020}), and lacked human validation studies.

We address these limitations by developing \textsc{SENSE}, projecting from the learned lexical embeddings of over 34k Lancaster words to their 11 human-annotated sensorimotor norm dimensions, and exploring these predictions through behavioral experiments with nearly 300 participants.
\section{Experiments}
We consider three hypotheses:

\vspace{-2pt}\paragraph{Hypothesis 1.} 
Lexical embeddings $\mathbf{e}_w$ of words $w$ implicitly encode sensorimotor norms $\mathbf{s}_w$, such that there exists a learned function ${f: \mathbb{R}^d \rightarrow \mathbb{R}^{11}}$ where ${f(\mathbf{e}_w) \approx \mathbf{s}_w}$ with lower error than a baseline predictor ${f_{\text{baseline}}(\mathbf{e}_w) = \bar{\mathbf{s}}}$, where ${\bar{\mathbf{s}} = \frac{1}{N}\sum_{i=1}^{N} \mathbf{s}_i}$.

\vspace{-2pt}\paragraph{Hypothesis 2.} 
Projection model $f$ captures systematic form-meaning correspondences that generalize sensorimotor associations to nonce words $w^*$ such that $f(\mathbf{e}_{w^*})_m \approx \mathbf{s}_{w^*}^m$ for modality $m$, where $\mathbf{s}_{w^*}^m$ is estimated with human responses.

\vspace{-2pt}\paragraph{Hypothesis 3.} 
Given \hyp{2} for modality $m$, character $n$-grams $c$ should exhibit systematic phonesthemic patterns where the human selection rate ${P_H(w^* \rightarrow m \mid c \in w^*)\propto f(\mathbf{e}_c)_m}$.

\vspace{3pt}We present the \textbf{S}ensorimotor \textbf{E}mbedding \textbf{N}orm \textbf{S}coring \textbf{E}ngine (\textsc{SENSE}), which learns the projection function $f$ (\hyp{1}). 
Then we conducted a human study to investigate whether $f$ can generalize its projected sensorimotor associations to nonce words (\hyp{2}), and whether those projections maintain correlation with human ratings at the character $n$-gram level (\hyp{3}).
All associated data and code can be found in \href{https://github.com/abhinav-usc/sense-model}{our code repository}.


\subsection{\textsc{SENSE} Projections}
The Sensorimotor Embedding Norm Scoring Engine (\textsc{SENSE}) projects lexical word embeddings onto the Lancaster Sensorimotor Norms.
\textsc{SENSE} takes in an embedding vector ${\mathbf{e}_w \in \mathbb{R}^d}$ and predicts sensorimotor norms ${f(\mathbf{e}_w) \in \mathbb{R}^{11}}$, where ${0 \leq f(\mathbf{e}_w)_m \leq 1}$ represents the predicted rating for word $w$ and modality $m$.\footnote{We normalize the Lancaster norms from their original $[0,5]$ range to $[0,1]$; note that the norms are not unit vectors.}

We selected words and phrases present in each of the Word2Vec, GloVe, and Lancaster Norms vocabularies, averaging constituent word vectors for multi-word phrases, leaving 34,110 aligned entries out of 39,707.
For the BERT model, lexical embeddings were obtained as the CLS representation of each word or phrase passed through the model alone without sentential context.
We randomly partitioned the selected words into training (70\%), development (15\%), and test (15\%) sets.

We compared three architectures: a baseline predicting mean training set sensorimotor vector ${\bar{\mathbf{s}}}$; $k$-NN with $k$=5 using cosine similarity with weighted averaging to predict $\mathbf{\hat{s}}_w$ from the neighborhood $\mathbf{s}_{NN(\mathbf{e}_w)}$; and a feed-forward neural network with one hidden layer, 64 or 128 neurons tuned on development set, ReLU activation, trained using Adam optimization with learning rate 0.001 for 10 epochs and batch size 128.
All models were evaluated using MSE on the held-out test set.

\subsection{Human Experiment}
To conduct experiments related to \hyp{2} and \hyp{3}, we selected the neural network architecture with BERT CLS embeddings as the \textsc{SENSE} model, enabling embedding of nonce words and arbitrary character sequences.

We conducted an IRB-approved human study surveying 281 undergraduate students about which nonce words evoked specific sensorimotor associations, allowing us to compare these annotations to \textsc{SENSE} predictions.
Nonce words were generated using the Wuggy Pseudoword Generator~\cite{Keuleers_Brysbaert_2010}, which creates pronounceable nonce words by preserving sub-syllabic structure and transition frequencies of real English words.
We used 63,975 seed words from Wuggy's lexicon, sourced from the 306,128-word Moby Word List~\cite{Ward_2002}, and generated 10 candidates per seed word using Wuggy's classic generator with a 2/3 sub-syllabic segment overlap ratio.

We retrieved BERT CLS embeddings for all generated nonce words and used \textsc{SENSE} to compute ${f(\mathbf{e}_{w^*})_m}$ for each nonce word $w^*$ and modality $m$.
To enforce lexical novelty, we excluded nonce words within Levenshtein distance 1 of, sharing stems with, or homophonous to Moby Dictionary entries.
We then selected the 12 nonce words with the highest ${f(\mathbf{e}_{w^*})_m}$ for each modality $m$, ensuring ${f(\mathbf{e}_{w^*})_m > 0.5}$ for all selections.
Table~\ref{tab:sample_generated_pseudowords} shows sample nonce words selected for each modality.

\begin{table}
    \centering
    \begin{tabular}{llr}
         \textbf{Nonce word $w^*$} & \textbf{Modality $m$} &
        \textbf{\textsc{SENSE} $\hat{s}^m_{w^*}$} \\
         \toprule
        crilollering    & Auditory       & $1.00$   \\
        caeduseousness  & Interoceptive  & $0.79$  \\
        rehotes         & Visual         & $0.90$  \\
        sweadles	    & Hand/Arm       & $0.75$  \\
        \bottomrule
    \end{tabular}
    \caption{Sample Wuggy nonce words with high corresponding modality norm predictions by \textsc{SENSE}.}
    \label{tab:sample_generated_pseudowords}
\end{table}

Using Qualtrics, we created a survey with 4 questions per modality asking: ``Which 3 of the following nonsense words do you think most relate to the [\sensory{modality}]?''
Each question presented 3 target words from the 12 selected for modality $m$ and 4 distractor words selected by ${f(\mathbf{e}_{w^*})_{m'}>0.5,m'\neq m}$ and with ${f(\mathbf{e}_{w^*})_m < 0.5}$. 
For example, one question asked: ``Which 3 of the following nonsense words do you think most relate to the sense of hearing?'' Two of its options were ``crilollering'' ($w_1^*$, $m=\sensory{auditory}$) and ``ancechuttos'' ($w_2^*$, $m'=\sensory{gustatory}$), where ${f(\mathbf{e}_{w_1^*})_{m} = 1}$, ${f(\mathbf{e}_{w_1^*})_{m'} \approx 0}$, ${f(\mathbf{e}_{w_2^*})_{m'} = 0.98}$, and ${f(\mathbf{e}_{w_2^*})_{m} \approx 0}$.

Each participant answered two questions per modality, and questions were balanced across participants.
The forced-choice design was chosen to simplify the task, as rating each nonce word independently across all 11 modalities would risk fatigue and yield noisy annotations given the inherent difficulty of judging novel words.
The 3/7 ratio of target nonce words per question yields a chance-level true positive rate of 42.9\%.
Since participants must always select three words regardless of perceived association strength, all results are interpreted against this chance baseline.

\subsection{Sublexical Analysis}
We investigated whether sublexical components $c$, character $n$-grams, increased the probability of a word evoking a particular sensory modality by computing $P_H(w^* \rightarrow m \mid c \in w^*)$ as the mean participant selection rate across all words containing $c$.
We aim to quantify the relationship between this selection rate and \textsc{SENSE}'s predictions.

The \sensory{interoceptive} and \sensory{auditory} modalities showed the strongest pseudoword correlations, so we decided to focus the sublexical analysis on these two modalities.
We extracted all character $n$-grams of length 2-4 from the 28 nonce words in each modality, retaining only character $n$-grams $c$ where $ | (w^*| c\in w^*) | \ \geq 3$ to ensure sufficient recurrence for statistical reliability.
To eliminate redundancy, we removed sub-string $n$-grams that appeared in the exact same word set as all their containing super-strings.
For each retained character set $c$ within modality $m$, we computed $P_H(w^* \rightarrow m \mid c \in w^*)$, eliminating character sets $c$ where $P_H \leq P_H(w^* \rightarrow m)$. 
This yielded 13 character sets for the \sensory{interoceptive} modality and 19 for \sensory{auditory}.
Table \ref{tab:sample_sublexical_data} presents some of the selected character sets $c$ for which $P_H - P(w^* \rightarrow m) > 0.10$.

\begin{table}[t]
    \centering
    \begin{tabular}{lrr}
         \textbf{Modality} & \textbf{Character Set ($c$)} & \textbf{$P_H$}
         \\
         \toprule
        \sensory{Auditory}       & in & $0.565$ \\
        \sensory{Auditory}       & er & $0.547$ \\
        \sensory{Interoceptive}  & ni & $0.579$ \\  
        \sensory{Interoceptive}  & tion & $0.538$  \\
        \bottomrule
    \end{tabular}
    \caption{Example sub-lexical components $c$ along with the probability $P_H = P_H(w^* \rightarrow m \mid c \in w^*)$ of a nonce word containing $c$ evoking the given sensory modality $m$.}
    \label{tab:sample_sublexical_data}
\end{table}

We tokenized each character set in order to obtain $\mathbf{e}_c$ and used \textsc{SENSE} to compute $f(\mathbf{e}_c)_m$. 
We then calculated Pearson correlations between ${P_H(w^* \rightarrow m \mid c \in w^*)}$ and $f(\mathbf{e}_c)_m$.

\section{Results}

\textsc{SENSE} results demonstrate that word embeddings encode sensorimotor information (\hyp{1}), the human study shows that \textsc{SENSE} predictions generalize to nonce words in alignment with human behavioral judgments (\hyp{2}), and sublexical analysis reveals systematic form-meaning associations in the \sensory{interoceptive} modality (\hyp{3}).

\subsection{\textsc{SENSE}}
For each sensorimotor modality $m_j$, we calculated $$\text{MSE}_{m_j} = {\frac{1}{N} \sum_{w_i=1}^N {(f(\mathbf{e}_{w_i})_{m_j}-\mathbf{s}_{w_i}^{m_j}})^2},$$ and evaluated overall performance $\text{MSE}_{\text{avg}}$ as the unweighted average MSE across all 11 modalities.

\begin{table}
  \centering
  \begin{tabular}{lrrr}
    \textbf{Embedding} & \textbf{Baseline} &
    \textbf{KNN} &
    \textbf{Neural Net} \\
    \hline
    \textbf{Word2Vec}     & {0.028} &{0.015} & {0.016}           \\
    \textbf{GloVe}  & {0.028} &{0.018} & {0.017}      \\
    \textbf{BERT CLS}     & {0.028} &{0.020} & {0.016}
               \\\hline
  \end{tabular}
  \caption{Overall performance of different model architectures (Baseline, KNN, Neural Network) $\text{MSE}_{\text{avg}}$, across 3 different embedding types (Word2Vec, GloVe, BERT CLS). For each sensorimotor modality $m_j$, we calculated $\text{MSE}_{m_j} = {\frac{1}{N} \sum_{w_i=1}^N {(f(\mathbf{e}_{w_i})_{m_j}-\mathbf{s}_{w_i}^{m_j}})^2}$ and evaluate $\text{MSE}_{\text{avg}}$ as the unweighted average MSE}
  
  \label{tab:proj_mod_results}
\end{table}
Table~\ref{tab:proj_mod_results} presents $\text{MSE}_{\text{avg}}$ across three embedding types $\textbf{e}$ (Word2Vec, GloVe, BERT CLS) and three architectures $f$ (Baseline, KNN, Neural Network). 
Both KNN and neural architectures substantially outperformed the baseline, demonstrating that learned projection functions can predict word-specific sensorimotor profiles rather than defaulting to average sensorimotor ratings based on their training data. 

\begin{figure}[t]
  \includegraphics[width=\columnwidth]{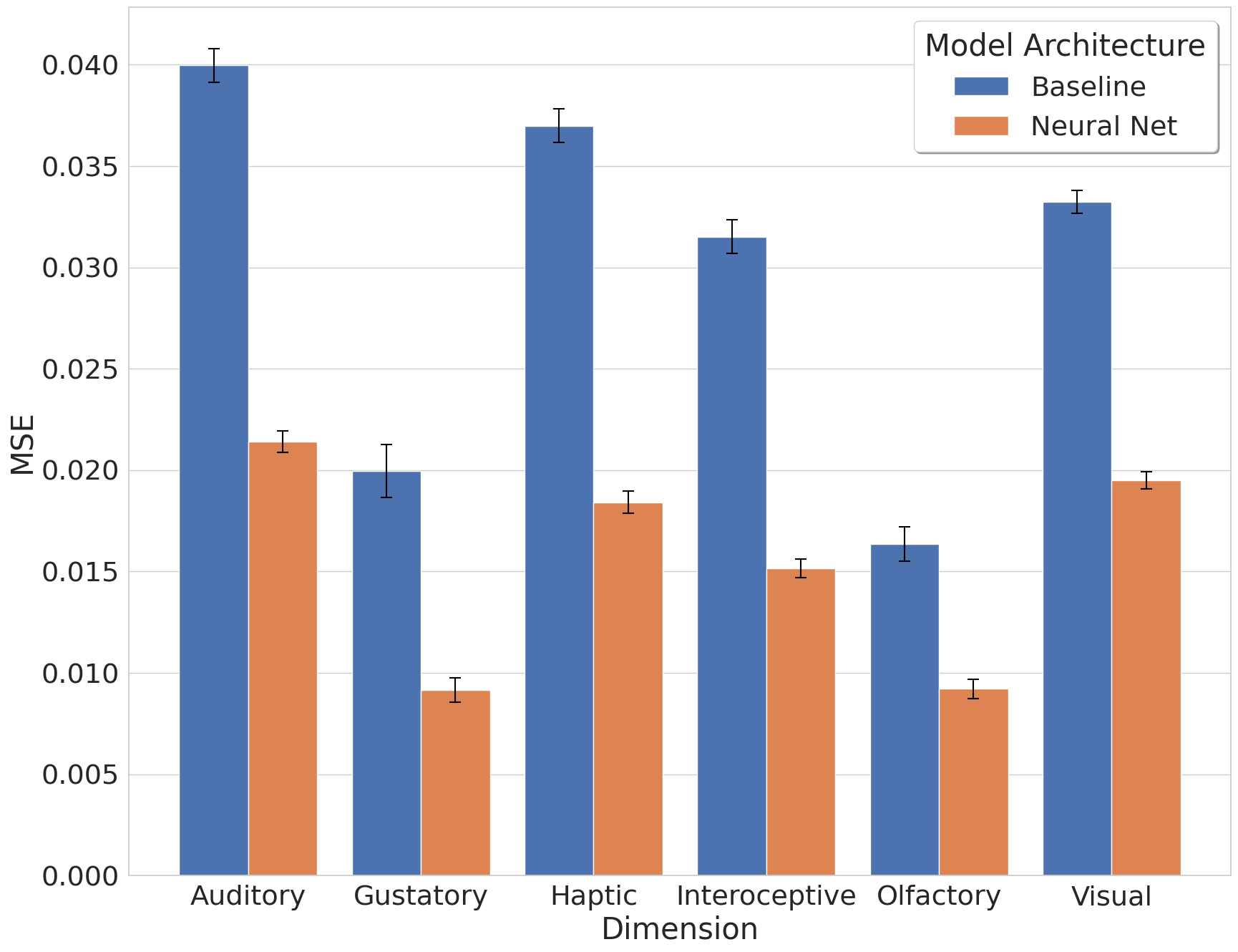}
  \vspace{-24pt}
  \caption{Mean squared error (MSE) for the six perceptual modalities comparing \textsc{SENSE} against the baseline model predicting mean training sensorimotor vector ${\bar{\mathbf{s}}}$ for all inputs. Error bars represent standard error of the MSE. 
  The neural network substantially outperforms the baseline across all modalities, with lowest errors for \sensory{gustatory} and \sensory{olfactory} and highest for \sensory{visual} and \sensory{auditory}.}
  \label{fig:comparision_modelling}
\end{figure}

Figure~\ref{fig:comparision_modelling} compares $\text{MSE}_{m_j}$ for the \textsc{SENSE} projections from BERT CLS versus the baseline average for the 6 perceptual modalities. 
Notably, the lowest error was observed for the \sensory{gustatory} and \sensory{olfactory} modalities across all three embeddings, indicating that taste and smell related concepts are well captured by word co-occurrence patterns. 
In contrast, \sensory{visual} and \sensory{auditory} dimensions showed the highest errors, but errors across all modalities were modest, suggesting that sensorimotor grounding is encoded in distributional semantics despite these models being trained solely on co-occurrence patterns. 

Paired $t$-tests comparing per-word MSE between \textsc{SENSE} and the baseline model revealed significantly lower errors for \textsc{SENSE} across all 11 modalities and overall ($p < .001$), providing strong support for \hyp{1}.



\subsection{Human Study}

For each nonce word $w^*$ and modality $m$, we computed the human selection rate (proportion of participants selecting it as one of the seven given nonce words that evoked sensorimotor modality $m$) and \textsc{SENSE}'s rating $f(\mathbf{e}_{w^*})_m$. 
Table \ref{tab:human_experiment_results} presents Pearson correlations between these measures for each modality that showed statistically significant correlations $r$ between human selection rate and \textsc{SENSE} predictions, sorted by $r$ value. 
Overall, \textsc{SENSE} predictions for 6 of the 11 modalities significantly correlated with human judgments, supporting \hyp{2}.
\begin{table}
    \centering
    \begin{tabular}{lr}
         \textbf{Modality} & \textbf{Correlation ($r$)} \\
         \toprule
         \sensory{Interoceptive} & $0.73^{***}$ \\
         \sensory{Auditory} & $0.69^{***}$ \\
         \sensory{Torso} & $0.57^{**\phantom{*}}$ \\
         \sensory{Visual} & $0.56^{**\phantom{*}}$ \\
         \sensory{Gustatory} & $0.54^{**\phantom{*}}$ \\
         \sensory{Hand/Arm} & $0.43^{*\phantom{**}}$ \\
        \bottomrule
    \end{tabular}
    \caption{Significant correlations between human selection rate and \textsc{SENSE} predictions, sorted by $r$ value. Five modalities (Foot/Leg, Olfactory, Haptic, Head, Mouth) showed non-significant correlations. $^{*}p < 0.05$, $^{**}p < 0.01$, $^{***}p < 0.001$.}
    \label{tab:human_experiment_results}
\end{table}

\begin{figure}[t]
  \includegraphics[width={\columnwidth}]{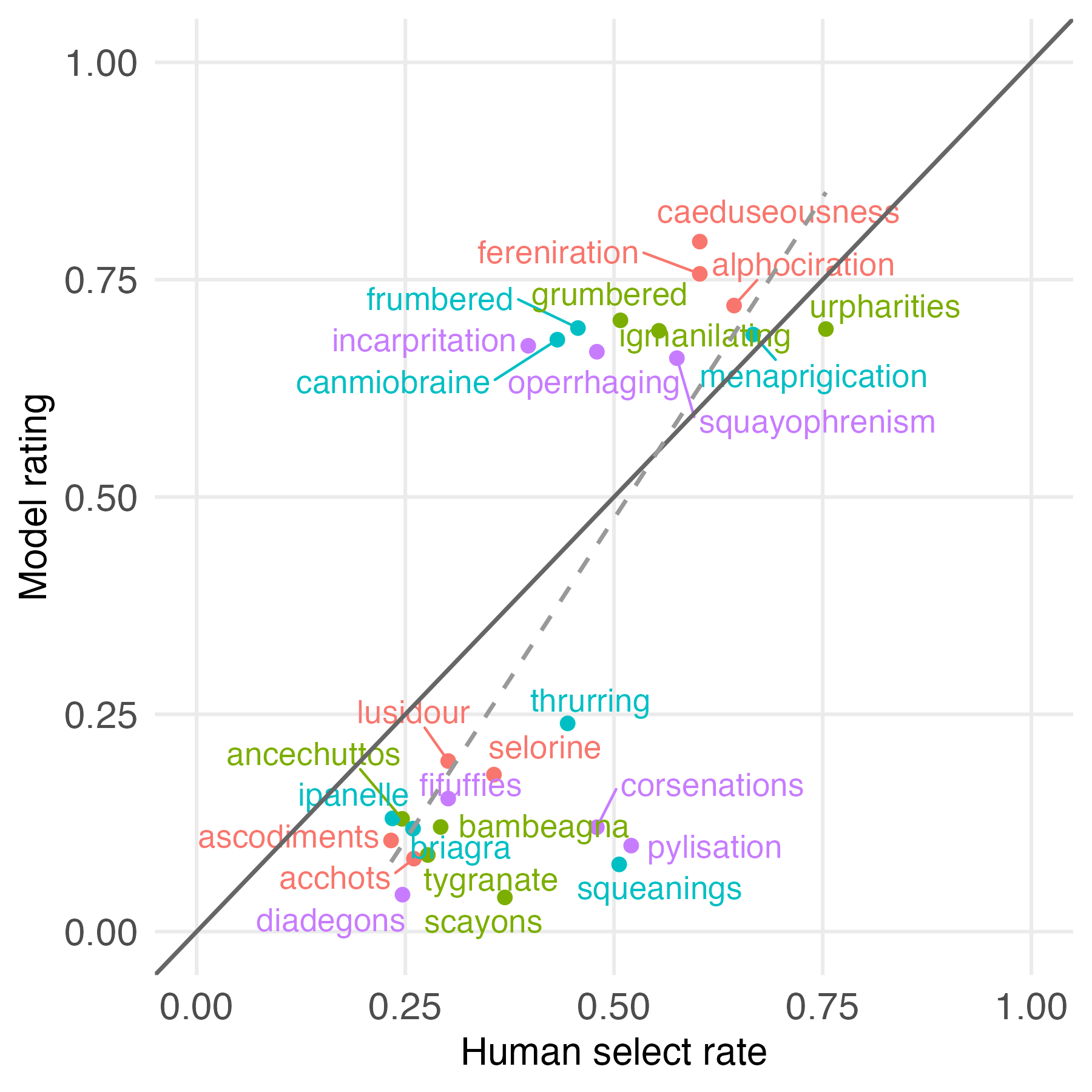}
  \vspace{-24pt}
  \caption{Correlation between the rate of human selection vs \textsc{SENSE}\ rating for nonce words shown to the humans under the \sensory{Interoceptive} category ($r=0.73$).}
  \label{fig:interoceptive_correlation}
\end{figure}

Figure \ref{fig:interoceptive_correlation} illustrates the correlation for words in the \sensory{interoceptive} dimension between human selection rate and \textsc{SENSE} ratings, showing that nonce words \textsc{SENSE} rated highly were also frequently selected by human participants, while low-rated nonce words were rarely selected.

\subsection{Sublexical analysis}
Since \sensory{interoceptive} and \sensory{auditory} modalities showed the strongest pseudoword correlations between human selection rates of nonce words and \textsc{SENSE}'s ratings ($r > 0.65$), we focused the sublexical analysis on these two modalities and found partial support for \hyp{3}.

\sensory{Interoception} showed a significant positive correlation between $P_H(w^* \rightarrow m \mid c \in w^*)$ and $f(\mathbf{e}_c)_m$ ($r = 0.630, p = 0.007$), indicating that character sets systematically associated with \sensory{interoceptive} experiences in human judgments are also captured in the model's embeddings.
However, \sensory{auditory} showed no significant relationship ($r = 0.267, p = 0.562$), suggesting that strong whole-word sensorimotor associations do not always decompose into systematic character-level phonesthemic patterns.

\section{Conclusion}

This work bridges distributional semantic models and embodied cognition theory, providing evidence of a relationship between text-based distributional representations and the sensorimotor grounding present in human language comprehension. 

We developed \textsc{SENSE}, a projection architecture that maps word embeddings onto the 11-dimensional Lancaster Sensorimotor Norms, demonstrating that distributional representations encode sensorimotor information with low prediction error across Word2Vec, GloVe, and BERT CLS embeddings. A behavioral study with 281 participants validated these predictions, showing strong correlations between human judgments and \textsc{SENSE} ratings for nonce words evoking specific sensorimotor experiences, with the strongest alignment in \sensory{interoception} ($r = 0.73, p < .001$). 

The absence of significant correlations for five modalities 
(\sensory{haptic}, \sensory{olfactory}, \sensory{foot/leg}, 
\sensory{head}, and \sensory{mouth/throat}) remains an open 
question. 
Preliminary analysis did not reveal clear correlations between modality performance and factors such as training corpus sparsity or inter-subject agreement, suggesting that this warrants 
dedicated future investigation. 
Sublexical analysis revealed systematic phonesthemic patterns for \sensory{interoceptive} experiences, suggesting the possibility of computational frameworks for proposing orthographic form-meaning associations.

\section*{Limitations}
This work has several limitations that suggest directions for future research.
A primary limitation of this work is that our evaluation is restricted to English, which constrains the cross-linguistic generalizability of our findings.
Since sensorimotor grounding and phonesthemes are often shaped by language-specific sound symbolism and orthographic conventions, we make no claim that SENSE would perform comparably in other languages. 
This is particularly relevant for morphologically rich or logographic languages, where the relationship between surface form and lexical semantics differs substantially from English.
Future work should cross-validate these findings with behavioral studies involving native speakers of typologically diverse languages, to better characterize how orthography, phonology, and distributional embeddings jointly encode sensorimotor information across linguistic systems.

Our behavioral validation relied on 281 undergraduate participants from a single university, introducing a WEIRD (Western, Educated, Industrialized, Rich, Democratic) sampling bias. 
Since sensorimotor experience is deeply tied to cultural and environmental context, it remains unclear how well the norms we captured generalize beyond our specific participant pool.

Additionally, our analysis examines orthographic forms rather than phonological representations.
Phonesthemes are traditionally phonological phenomena, yet our analysis relies on written text due to the text-only training data underlying word embeddings and language models.
The form-meaning associations we detect might be stronger with phonological representations that capture actual sound patterns.

Furthermore, we used BERT to generate static, context-free word embeddings, which does not fully leverage its representational capacity, as contextualized embeddings may carry richer sensorimotor information. 
However, this choice was deliberate: our human study centers on nonce words, which lack naturalistic sentential contexts. 
Using CLS-token representations of isolated words minimizes the train-to-inference domain mismatch that would arise if we projected nonce word embeddings into a space trained on average contextualized representations.

Finally, while our sublexical analysis revealed systematic patterns for interoceptive phonesthemes, time constraints prevented a targeted behavioral study to validate these form-meaning correspondences directly. Such a study would present participants with words containing character sets selected based on the model's predicted interoceptive associations, providing a direct test of whether these patterns are psychologically real. This remains a natural and important direction for future work.

\section*{Acknowledgments}
This work was supported in part by a USC Undergraduate Research Associates Program (URAP) grant to Thomason, and a USC Provost's Fellowship and the Daben Weiqing Liu Research Fellowship awarded to Gupta.
\bibliography{anthology,custom}

\clearpage
\appendix

\section{Sample Lancaster Words and their sensorimotor ratings}
\label{sec:samples}

\begin{figure} 
  \centering
  \begin{minipage}{0.48\columnwidth}
    \includegraphics[width=\textwidth]{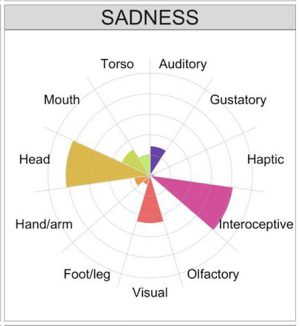}
  \end{minipage}
  \hfill
  \begin{minipage}{0.48\columnwidth}
    \includegraphics[width=\textwidth]{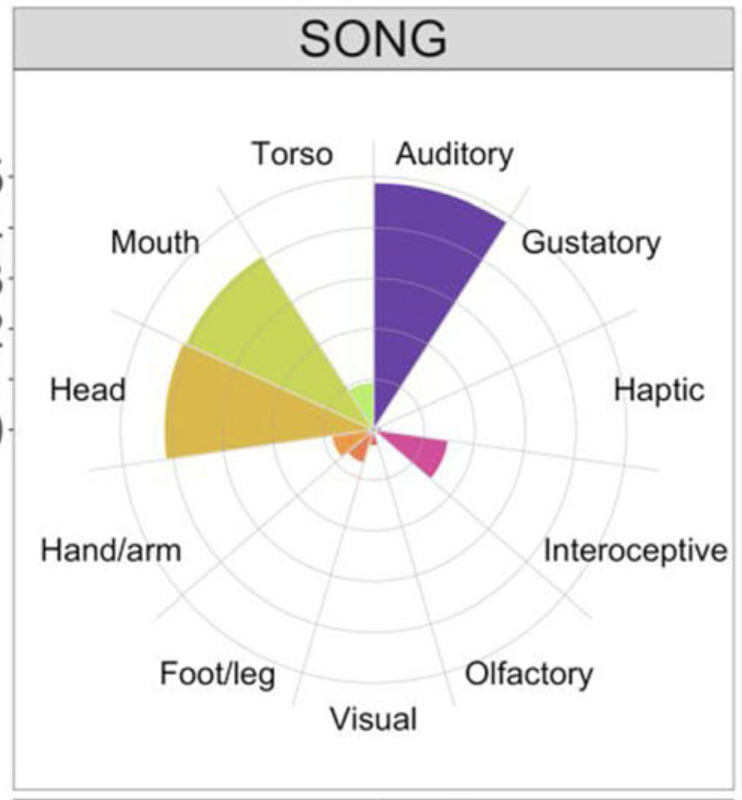}
  \end{minipage}
  
  \vspace{0.5em}
  
  \begin{minipage}{0.48\columnwidth}
    \includegraphics[width=\textwidth]{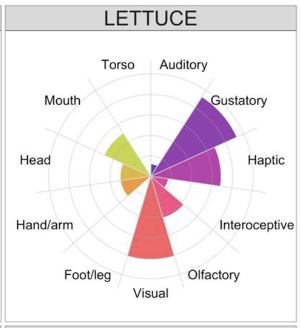}
  \end{minipage}
  \hfill
  \begin{minipage}{0.48\columnwidth}
    \includegraphics[width=\textwidth]{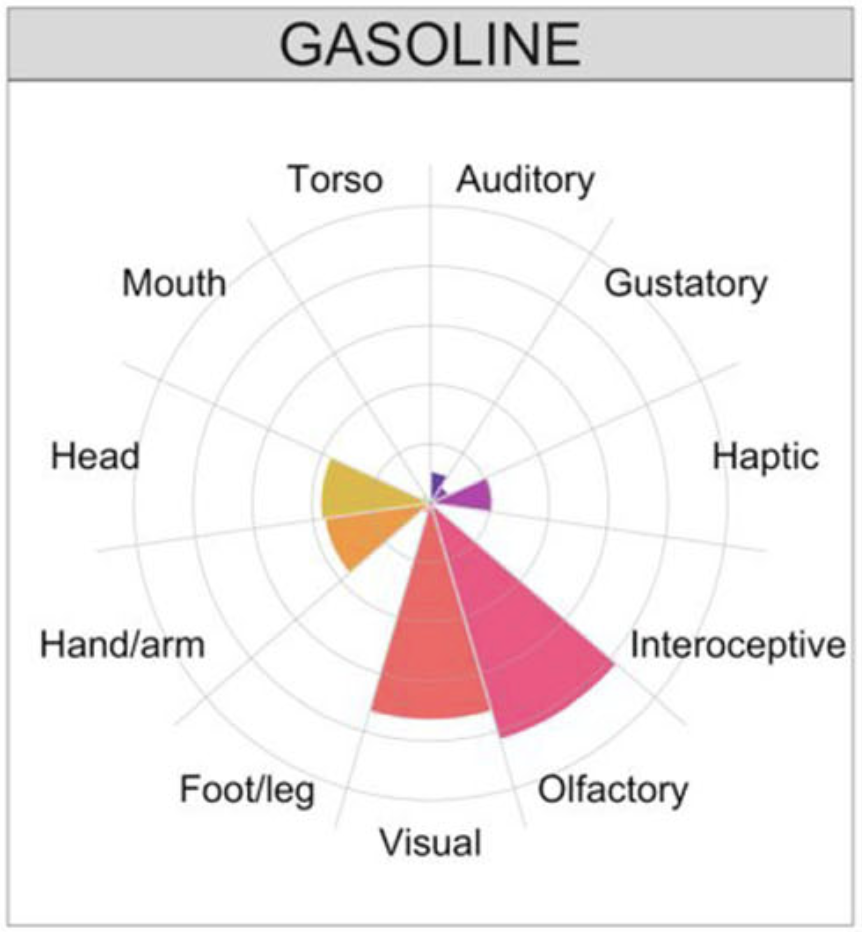}
  \end{minipage}
  
  \caption{Sample words from the Lancaster Sensorimotor Norms dataset showing ratings across 11 dimensions (6 perceptual modalities and 5 action effectors).}
  \label{fig:lanc_examples}
\end{figure}

Figure \ref{fig:lanc_examples} contains examples from the Lancaster Sensorimotor Norms dataset, showing how words are rated across 11 sensorimotor dimensions: Auditory, Gustatory, Haptic, Interoceptive, Olfactory, Visual (perceptual), and Foot/leg, Hand/arm, Head, Mouth, Torso (action effectors).

\newpage
\section{Per-Modality Correlation Plots}
\label{sec:correlations}

\begin{figure} 
  \includegraphics[width={0.9\columnwidth}]{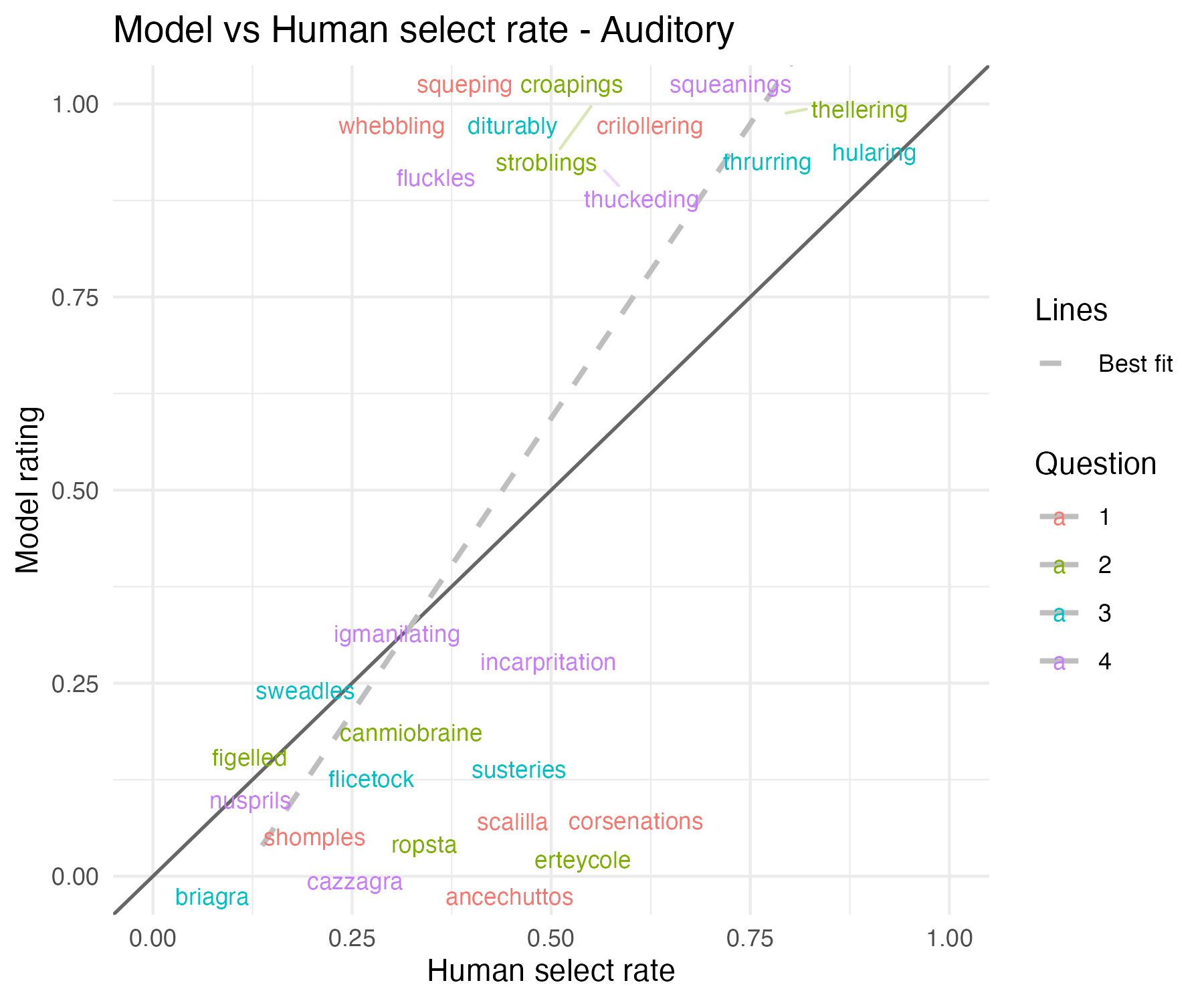}
  \caption{Correlation between human selection rate and \textsc{SENSE} ratings for nonce words in the \sensory{auditory} modality ($r = 0.69, p < .001$).}
  \label{fig:auditory_correlation}
\end{figure}

\begin{figure} 
  \includegraphics[width={0.9\columnwidth}]{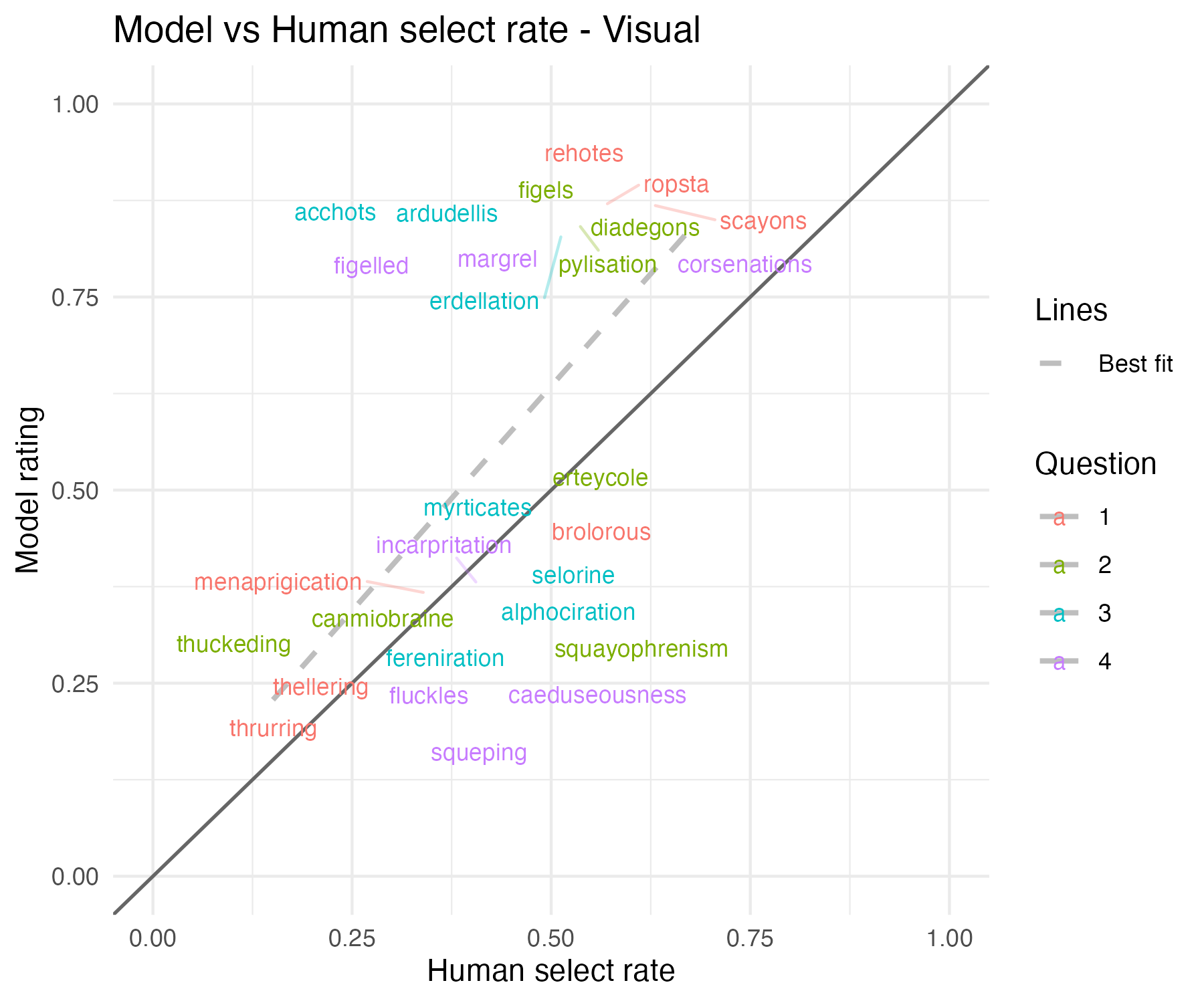}
  \caption{Correlation between human selection rate and \textsc{SENSE} ratings for nonce words in the \sensory{visual} modality ($r = 0.56, p = .002$).}
  \label{fig:visual_correlation}
\end{figure}

\begin{figure} 
  \includegraphics[width={0.9\columnwidth}]{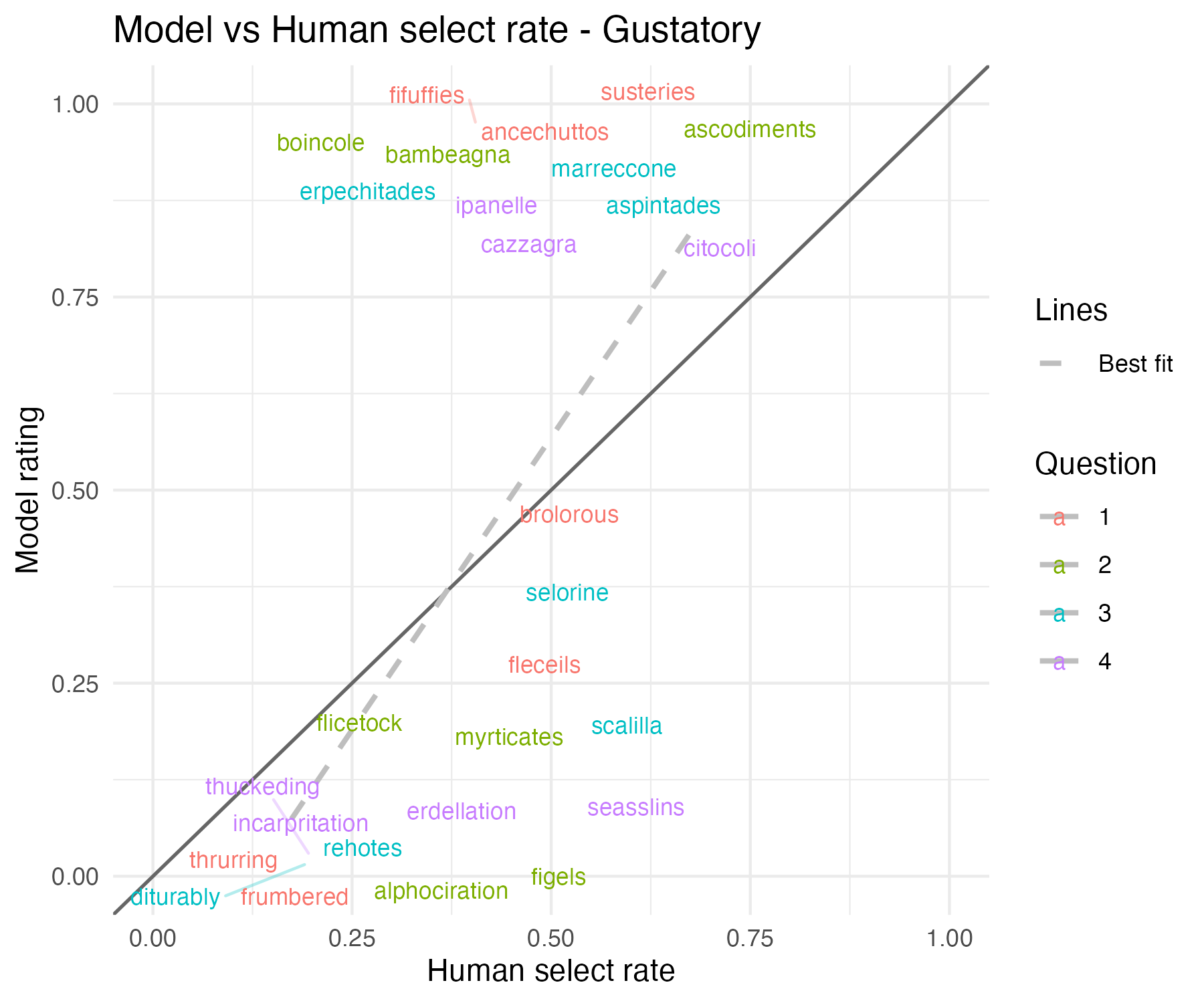}
  \caption{Correlation between human selection rate and \textsc{SENSE} ratings for nonce words in the \sensory{gustatory} modality ($r = 0.54, p = .003$).}
  \label{fig:gustatory_correlation}
\end{figure}

\begin{figure} 
  \includegraphics[width={0.9\columnwidth}]{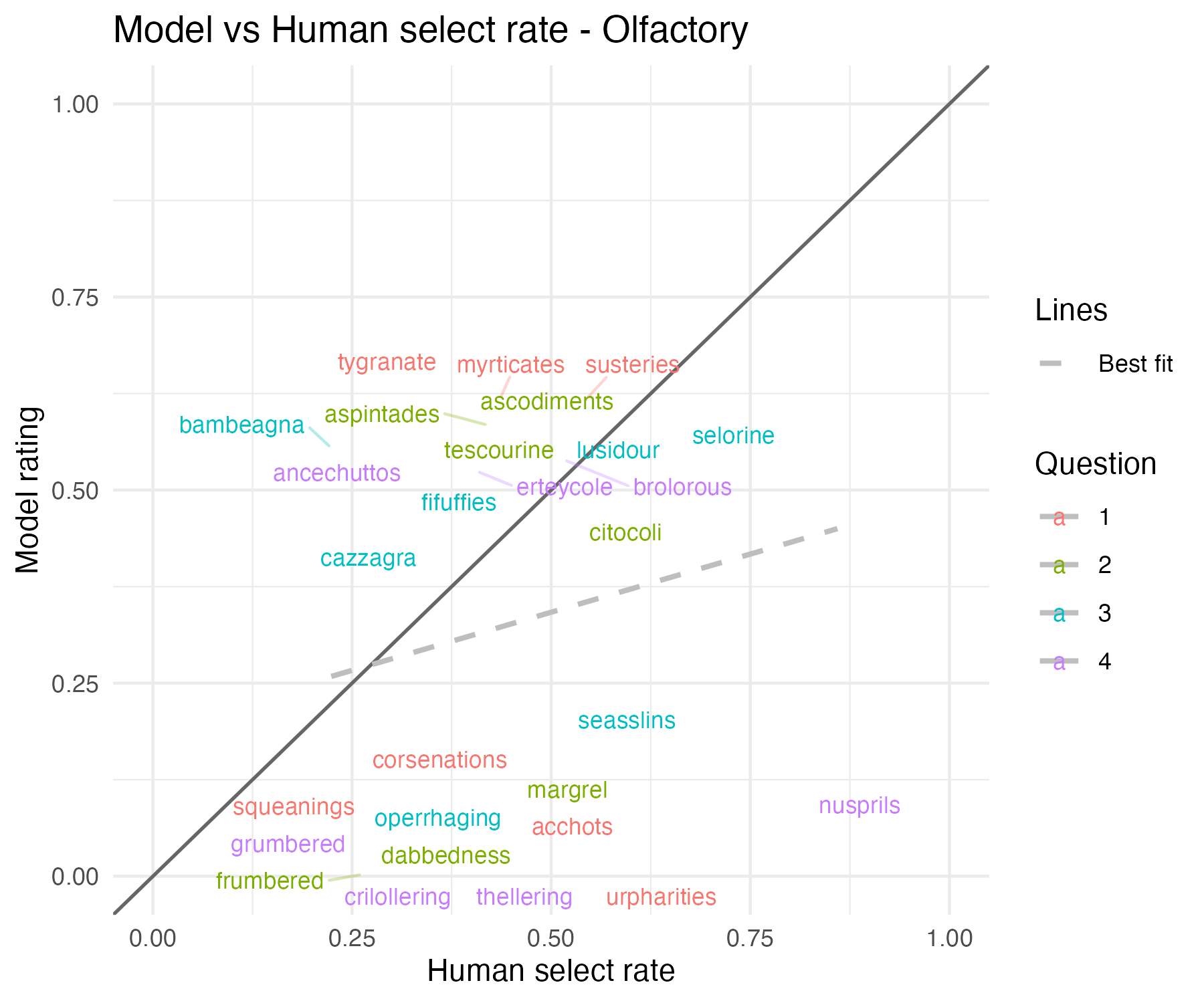}
  \caption{Correlation between human selection rate and \textsc{SENSE} ratings for nonce words in the \sensory{olfactory} modality ($r = 0.18, p = .349$).}
  \label{fig:olfactory_correlation}
\end{figure}

\begin{figure} 
  \includegraphics[width={0.9\columnwidth}]{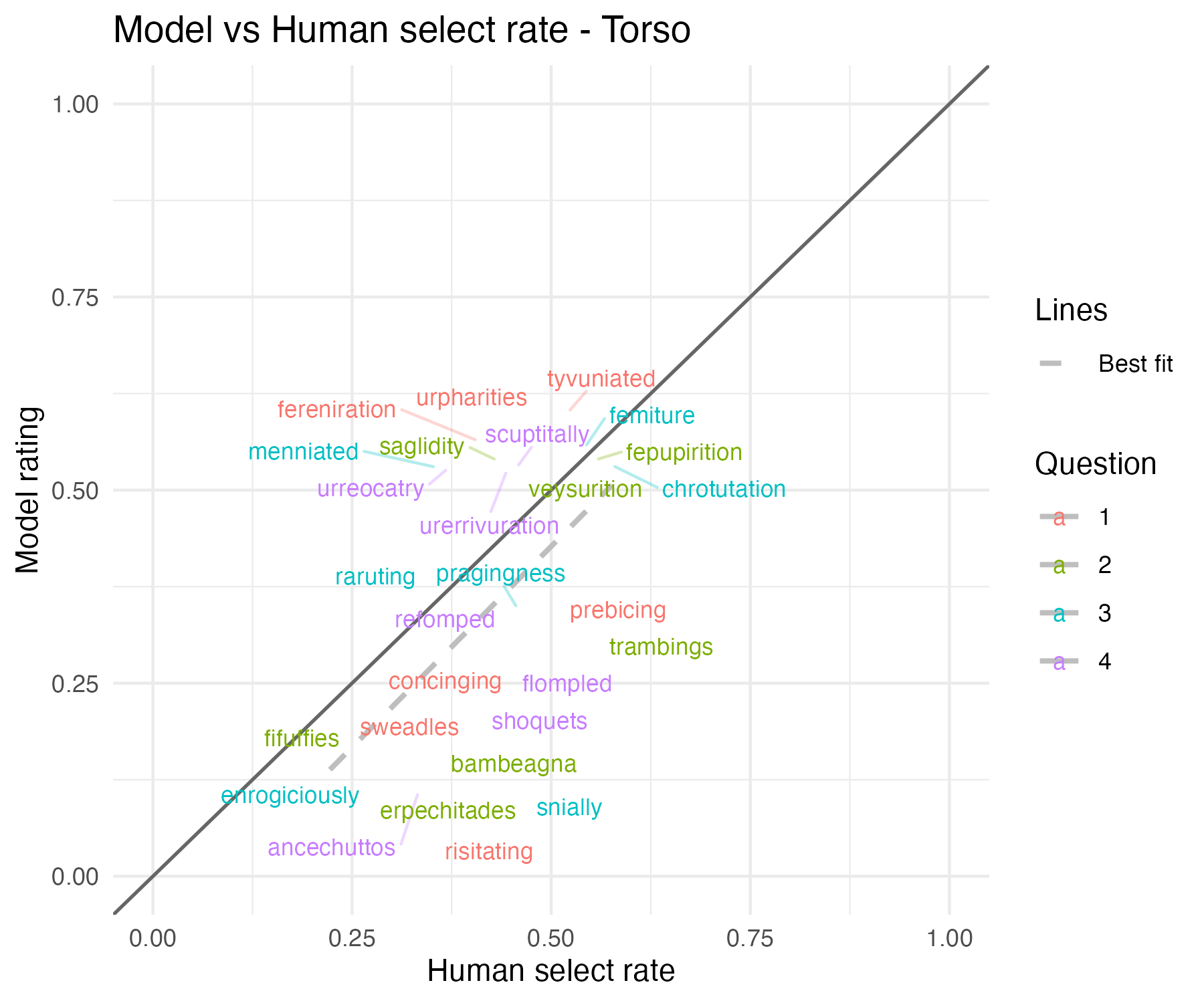}
  \caption{Correlation between human selection rate and \textsc{SENSE} ratings for nonce words in the \sensory{torso} modality ($r = 0.57, p = .002$).}
  \label{fig:torso_correlation}
\end{figure}

\begin{figure} 
  \includegraphics[width={0.9\columnwidth}]{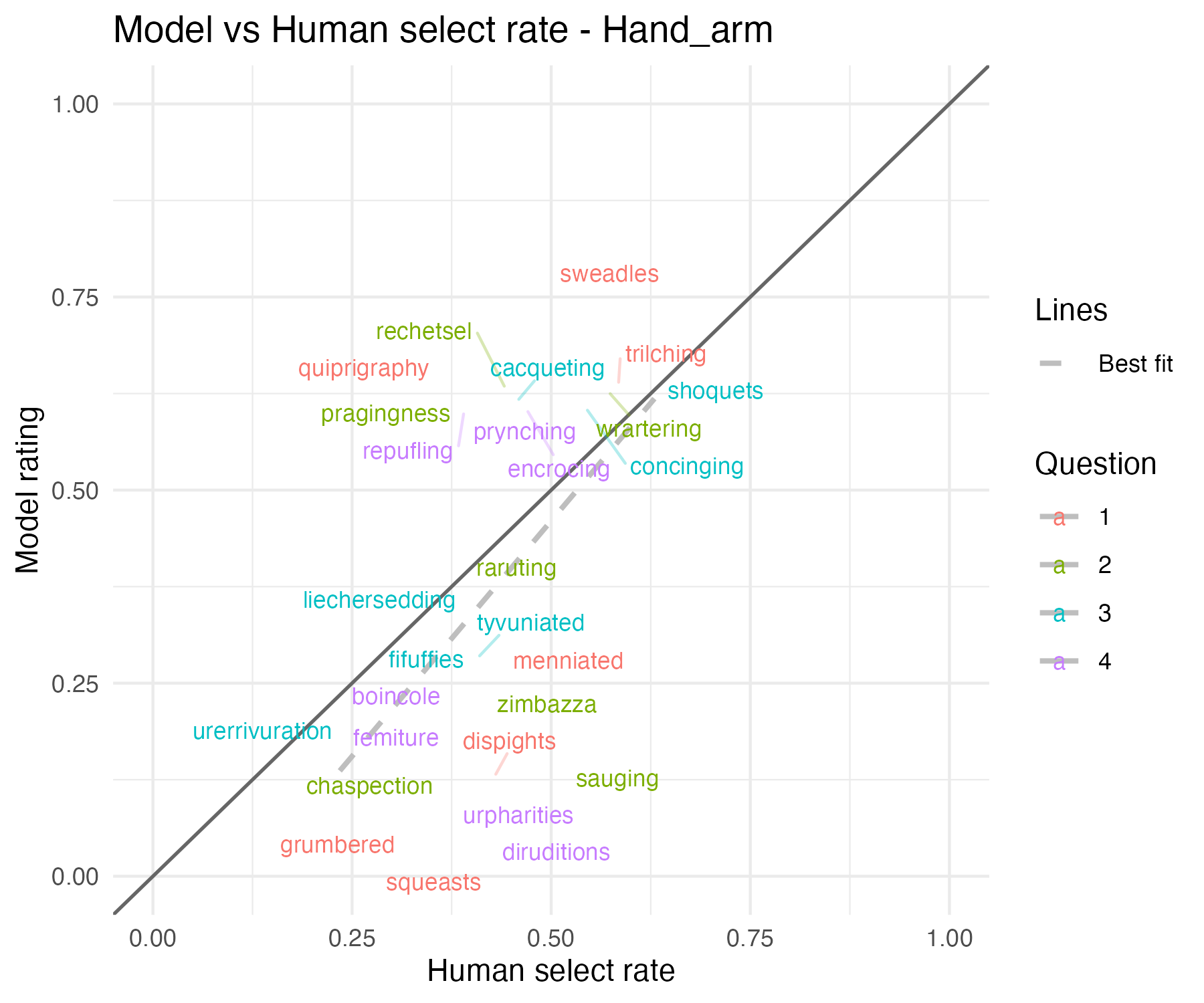}
  \caption{Correlation between human selection rate and \textsc{SENSE} ratings for nonce words in the \sensory{hand/arm} modality ($r = 0.43, p = .021$).}
  \label{fig:handarm_correlation}
\end{figure}

\begin{figure} 
  \includegraphics[width={0.9\columnwidth}]{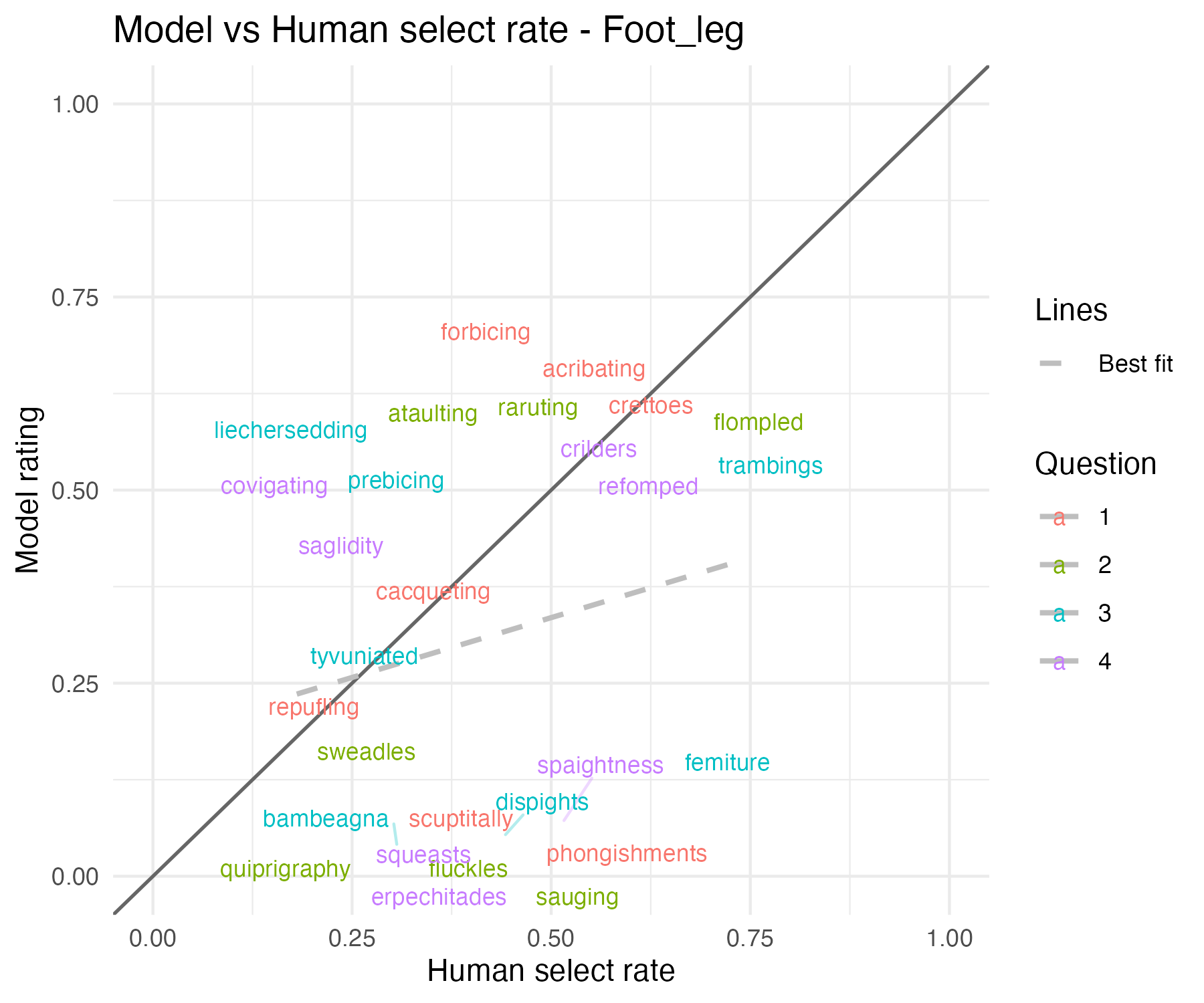}
  \caption{Correlation between human selection rate and \textsc{SENSE} ratings for nonce words in the \sensory{foot/leg} modality ($r = 0.25, p = .192$).}
  \label{fig:footleg_correlation}
\end{figure}

\begin{figure} 
  \includegraphics[width={0.9\columnwidth}]{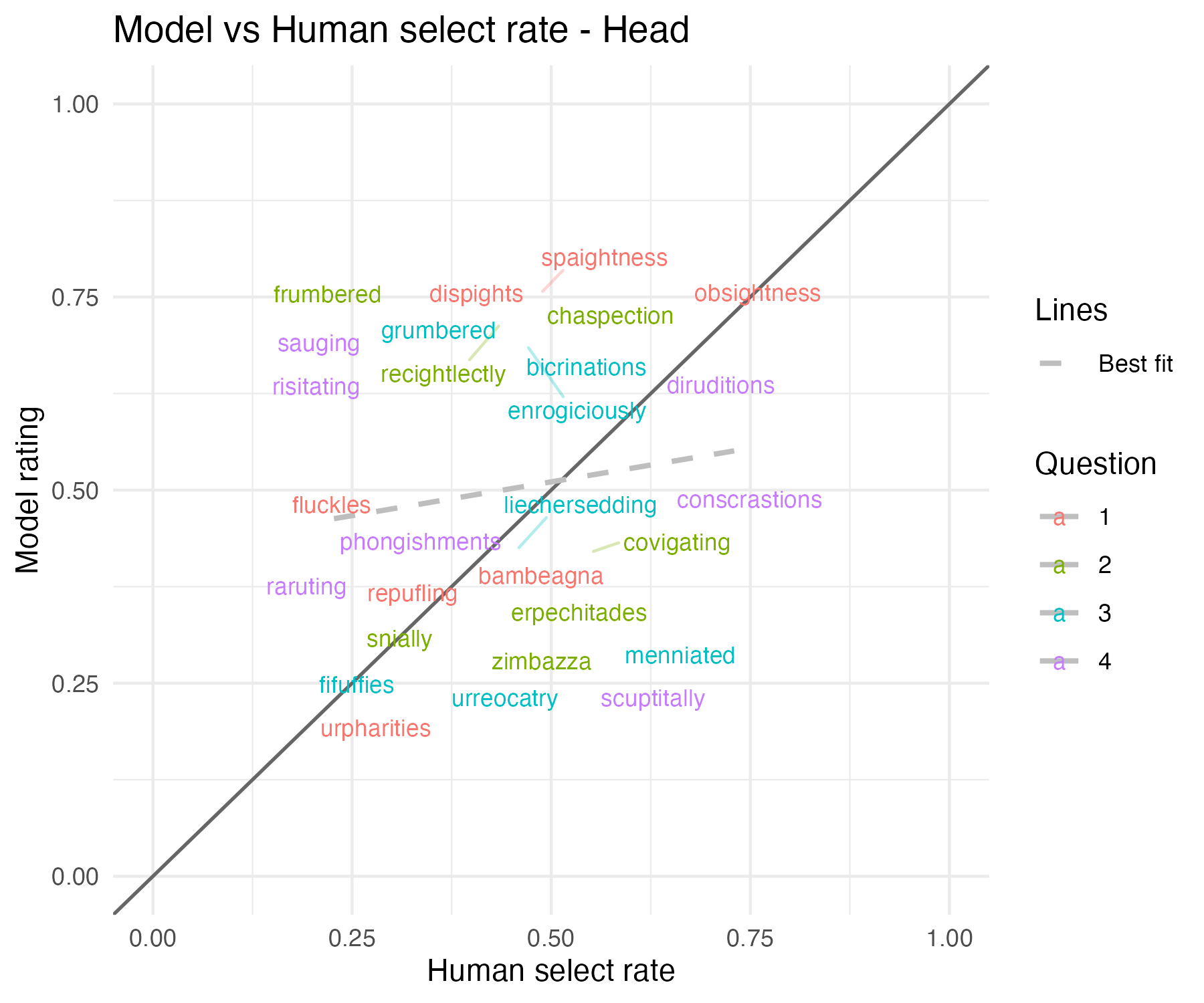}
  \caption{Correlation between human selection rate and \textsc{SENSE} ratings for nonce words in the \sensory{head} modality ($r = 0.10, p = .618$).}
  \label{fig:head_correlation}
\end{figure}

\begin{figure} 
  \includegraphics[width={0.9\columnwidth}]{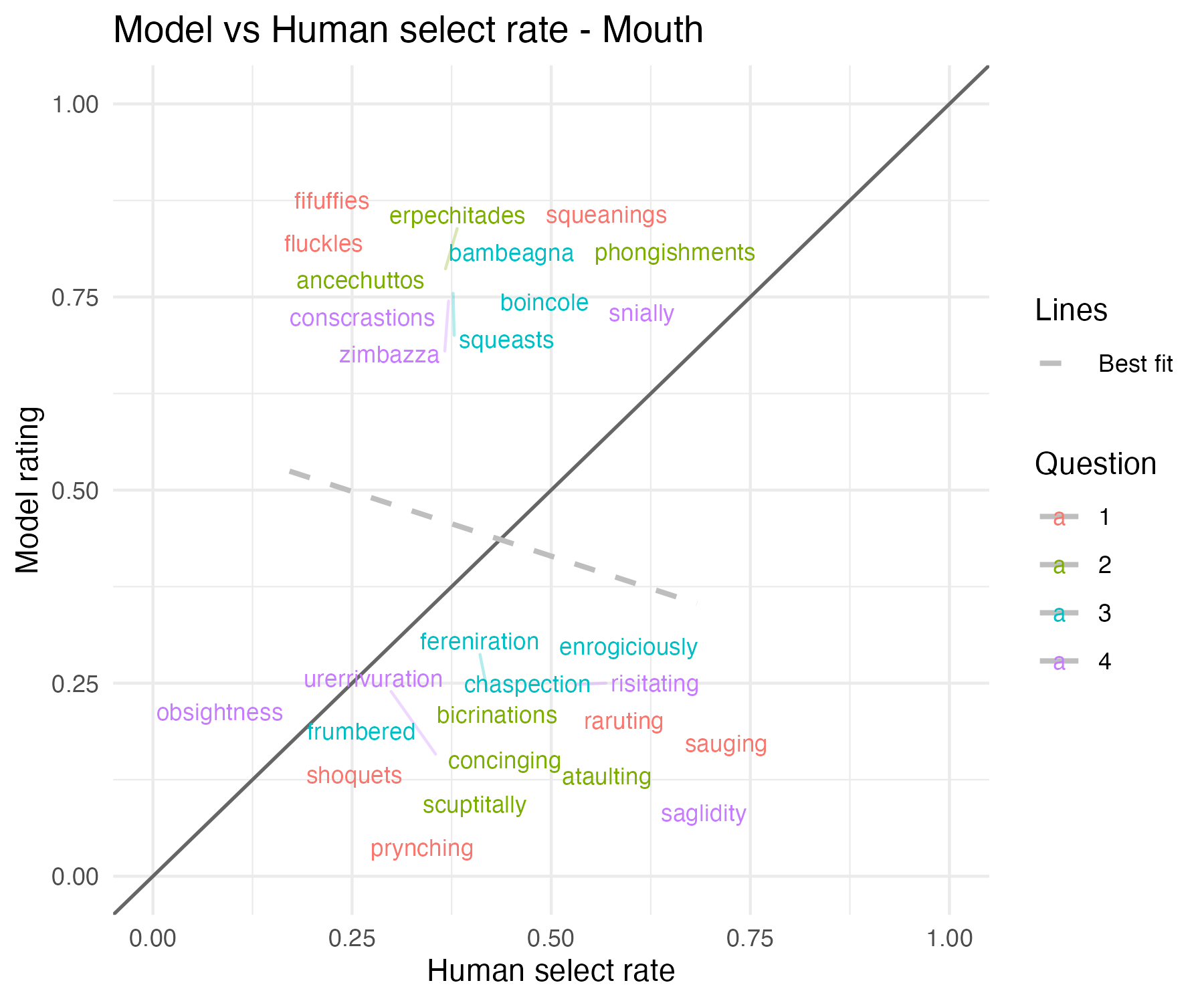}
  \caption{Correlation between human selection rate and \textsc{SENSE} ratings for nonce words in the \sensory{mouth} modality ($r = -0.14, p = .474$).}
  \label{fig:mouth_correlation}
\end{figure}

\begin{figure} 
  \includegraphics[width={0.9\columnwidth}]{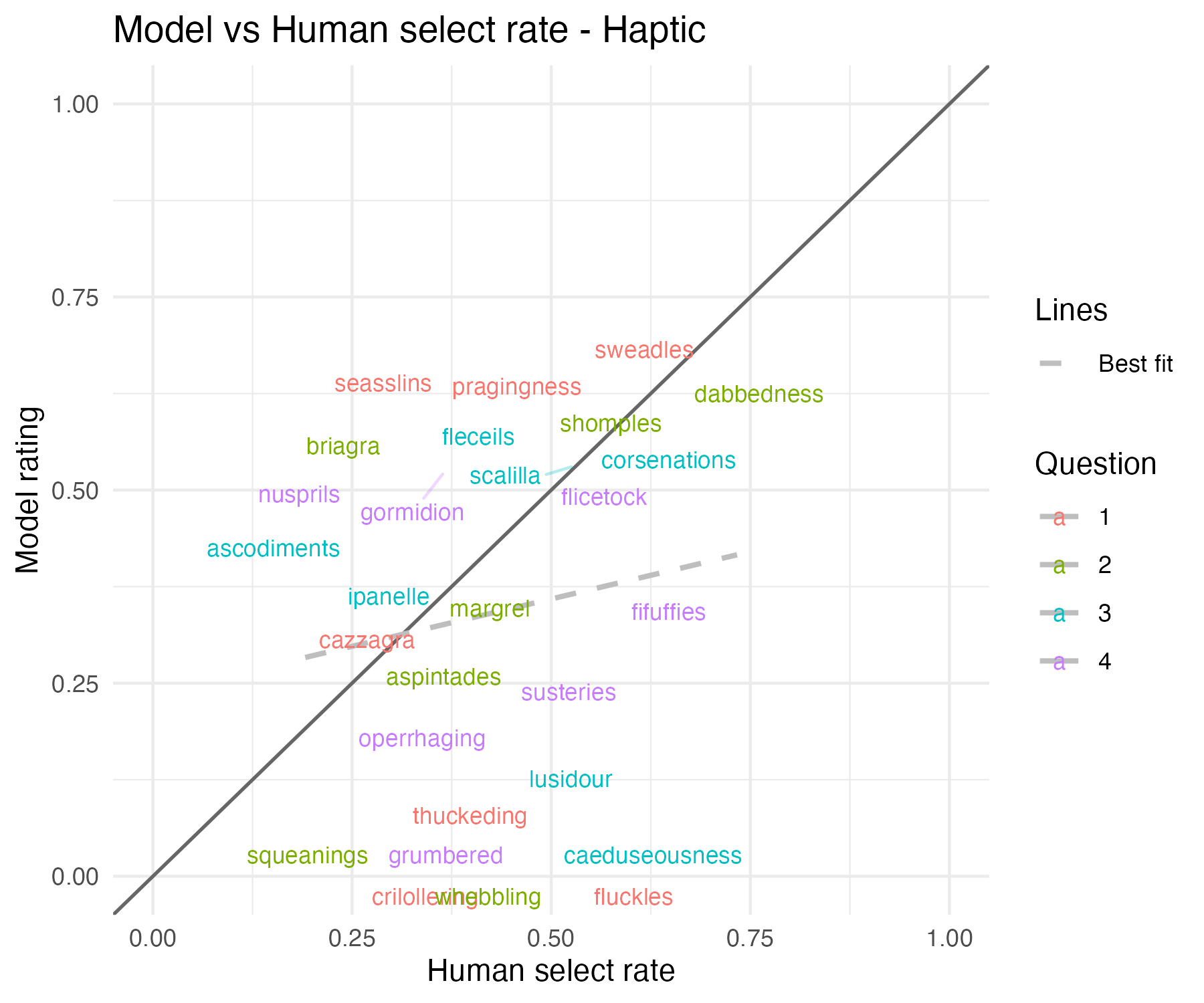}
  \caption{Correlation between human selection rate and \textsc{SENSE} ratings for nonce words in the \sensory{haptic} modality ($r = 0.16, p = .408$).}
  \label{fig:haptic_correlation}
\end{figure}

Figures \ref{fig:auditory_correlation}--\ref{fig:haptic_correlation} show correlations between human selection rates and \textsc{SENSE} ratings for the 10 remaining modalities tested in the behavioral study.

\end{document}